\documentclass[runningheads]{llncs}

\usepackage{eccv}

\usepackage{eccvabbrv}

\usepackage{graphicx}
\usepackage{booktabs}
\usepackage{multirow}
\usepackage{amsmath}   %
\usepackage{xcolor}    %
\usepackage{subcaption}

\usepackage{soul}

\usepackage[accsupp]{axessibility}  %

\usepackage[pagebackref,breaklinks,colorlinks,citecolor=eccvblue]{hyperref}
\usepackage{hyperref}

\usepackage{orcidlink}

\usepackage{wrapfig}

\usepackage{pifont}

\begin{document}

\title{DynFlowDrive: Flow-Based Dynamic World Modeling for Autonomous Driving}

\author{Xiaolu Liu\inst{1,2} \and
Yicong Li\inst{2}\and
Song Wang\inst{1} \and
Junbo Chen\inst{3}* \and \\
Angela Yao\inst{2} \and
Jianke Zhu\inst{1}\thanks{Corresponding authors.}}

\authorrunning{F.~Author et al.}

\institute{$^{1}$Zhejiang University, 
$^{2}$National University of Singapore, $^{3}$Udeer.AI \\
\{xialuliu, jkzhu\}@zju.edu.cn
}

\authorrunning{X. Liu et al.}
\titlerunning{DynFlowDrive}
\maketitle

 \vspace{-4mm}
\begin{figure}[!ht]
    \centering
    \includegraphics[width=\textwidth]{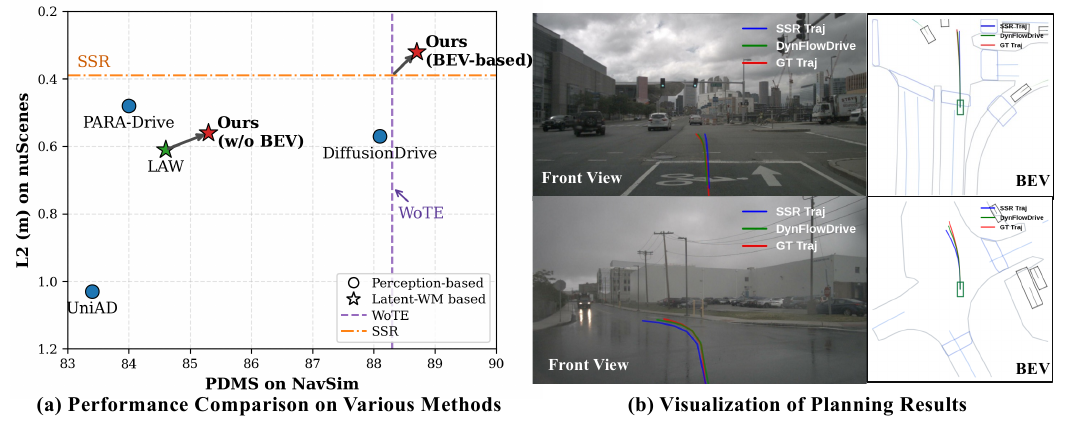}
   \vspace{-6mm}
    \caption{\textbf{(a)} Comparisons of perception-based and latent world model-based approaches on nuScenes and NavSim benchmarks. \textbf{(b)} Planning visualization on the front view and bird's-eye-view (BEV) space. Our DynFlowDrive achieves comparable performance.}
    \vspace{-8mm}
    \label{fig:teaser} %
\end{figure}

\begin{abstract}

Recently, world models have been incorporated into the autonomous driving systems to improve the planning reliability. Existing approaches typically predict future states through appearance generation or deterministic regression, which limits their ability to capture trajectory-conditioned scene evolution and leads to unreliable action planning.
To address this, we propose \textbf{DynFlowDrive}, a latent world model that leverages flow-based dynamics to
model the transition of world states under different driving actions. By adopting the rectified-flow formulation, the model learns a velocity field that describes how the scene state changes under different driving actions, enabling progressive prediction of future latent states.
Building upon this, we further introduce a stability-aware multi-mode trajectory selection strategy that evaluates candidate trajectories according to the stability of the induced scene transitions.
Extensive experiments on the nuScenes and NavSim benchmarks demonstrate consistent improvements across diverse driving frameworks without introducing additional inference overhead. Source code will be available at \href{https://github.com/xiaolul2/DynFlowDrive}{https://github.com/xiaolul2/DynFlowDrive}.

\vspace{-2mm}

\end{abstract}

\section{Introduction}
End-to-end autonomous driving has emerged as a promising paradigm for building driving systems~\cite{chen2024end,chib2023recent, dong2025end,hu2025vision}. Given the data captured by onboard sensors, it aims to predict safe and reliable future trajectories for planning and control~\cite{hu2025vision, chen2024vadv2, weng2024paradrive}. However, trajectory planning is a highly interactive process with the surroundings. A planned trajectory represents the intended action whose safety depends on how the environment responds after it is executed. 
Therefore, achieving reliable and safe planning requires anticipating future scene evolution, which remains a key challenge for autonomous driving systems.%

To equip the vehicle with such foresight capabilities, a new line of research introduces world models~\cite{allen1983planning, kong20253d} for the end-to-end methods~\cite{guan2024world, wang2024drivingfuture}. By enabling action-conditioned future simulations, these approaches allow the system to foresee the consequences of candidate trajectories and anticipate potential risks, thereby supporting safer and more reliable driving decisions. 
World models mainly follow two approaches: one line of work predicts future scenes explicitly, for example, by generating pixel-level images or 3D occupancy grids~\cite{wang2024drivedreamer, zhang2025epona, zheng2024occworld, yang2025drivingoccworld}. In contrast, another line of work models the scene prediction of the world in a latent space~\cite{li2024law, zheng2025world4drive, yang2025worldrft}, without explicitly reconstructing the full scenes.

Despite the advances, existing world models still suffer from critical limitations. For explicit scene-level prediction, many approaches rely on diffusion~\cite{podell2023sdxl,gao2023magicdrive} or autoregressive models~\cite{zhang2025epona,hu2023gaia} to synthesize future observations. While produced visually pleasing results, these methods primarily focus on appearance details such as textures and lighting. The emphasis on high-frequency visual patterns overlooks the underlying geometry and dynamics, which introduces extra computational overhead without improving action-oriented reasoning.

\begin{figure}[!ht]
    \centering
    \includegraphics[width=\textwidth]{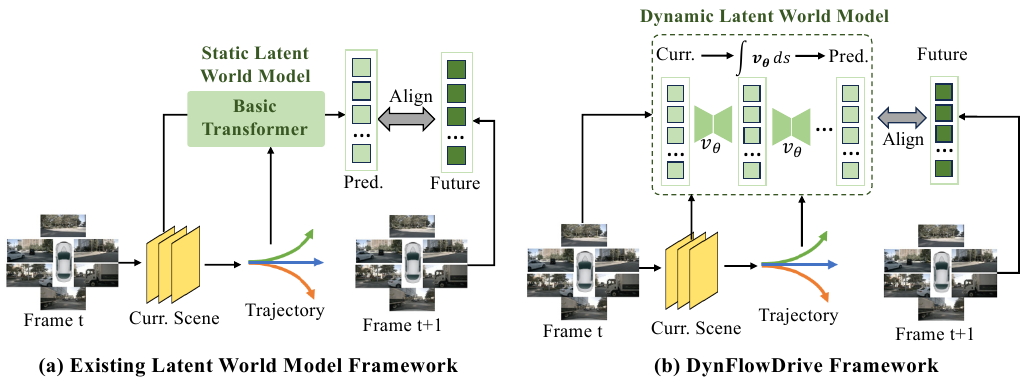}
    \vspace{-6mm}
    \caption{Comparison between \textbf{(a)} the existing \textbf{static world model} and \textbf{(b)} the \textbf{dynamic latent world model of our DynFlowDrive}. Instead of the static regression of next-frame latents, we propose the dynamic modeling that learns a continuous velocity field $v_{\theta}$ to capture the evolution of world transitions.}
    \vspace{-7mm}
    \label{fig:compare} %
\end{figure}

On the other hand, latent-space world models shift the focus from appearance synthesis to inner 
features
~\cite{li2024law, zheng2025world4drive}. 
As illustrated in Figure~\ref{fig:compare} (a), most works adopt the one-step regression that directly maps the current latent state to the next timestep. Such a static formulation simplifies the prediction as a rigid mapping and neglects the dynamic transition process, making it hard to assess the safety and feasibility of candidate actions. %
For instance, when a vehicle approaches pedestrians, it may either gradually slow down to yield or maintain speed and brake abruptly at the last moment. 
Although both actions may lead to a similar stopping position, the underlying scene evolution and safety implications are fundamentally different. 
Without modeling this transition process, the world model cannot faithfully capture trajectory-dependent dynamics or assess whether the induced evolution is smooth and physically plausible.

To bridge this gap, we argue that effective world modeling must move beyond discrete endpoint regression and explicitly formulate trajectory-conditioned scene evolution as a continuous dynamical system.
Motivated by this, we propose \textbf{DynFlowDrive}, a latent world model
that leverages flow-based dynamics to model the transition of world states under different driving actions. Specifically, based on the extracted latent features, 
DynFlowDrive adopts a rectified-flow formulation to model the transition dynamics in latent space and predict future latent states conditioned on the current observation and multiple candidate trajectories. The learned velocity field explicitly captures the rate of change of the scene during state transitions, enabling the model to capture how the environment progressively evolves under different driving actions.
Building upon the learned flow dynamics, we further introduce the stability-aware multi-mode trajectory selection mechanism. By exploiting the velocity field produced by the flow-based world model, we derive a stability measure that reflects the smoothness of trajectory-conditioned scene transitions. 
Combined with trajectory error measured with the ground truth and latent reconstruction discrepancy, this dynamics-aware stable criterion enables more reliable planning assessment and trajectory selection for safer autonomous driving systems. %

For evaluation, DynFlowDrive achieves comparable performance on challenging benchmarks, including nuScenes~\cite{caesar2020nuscenes} and NavSim~\cite{dauner2024navsim}.
As shown in Figure~\ref{fig:teaser}, our method consistently improves planning accuracy for BEV-based and w/o BEV approaches. Without auxiliary tasks, DynFlowDrive is still comparable to perception-based models. In particular, compared with SSR~\cite{li2024ssr}, DynFlowDrive reduces the average $L_2$ displacement error from 0.39m to 0.31m and achieves 88.7 PDMS on NavSim.
Our main contributions are summarized as follows:
\vspace{-1mm}
\begin{itemize}
    \item We propose \textbf{DynFlowDrive}, a dynamic latent world model that leverages rectified flow to explicitly model dynamic world evolution, enabling the progressive next-frame latent prediction conditioned on candidate trajectories.
    \item We introduce the stability-aware multi-mode selection strategy that leverages the flow-based stability for more reliable trajectory selection.
    \item Extensive experiments demonstrate the consistent and stable performance improvements across autonomous driving paradigms, without introducing additional computational overhead during inference.
\end{itemize}

\section{Related Work}
\noindent{\textbf{End-to-End Autonomous Driving.}}
Conventional autonomous systems decompose the whole paradigm into three stages:  perception~\cite{li2024bevformer,wang2024not,wang2023exploring,liu2024mgmap,liu2025uncertainty}, prediction~\cite{song2020pip,xin2018intention,liu2021survey}, and planning~\cite{lim2019hybrid,gonzalez2015review}. Such modular designs often lead to error accumulation and substantial engineering overhead. 
To address these limitations, recent works explore end-to-end autonomous driving, which learns a direct mapping from sensory observations to driving actions or trajectories~\cite{hu2023planning,sun2025sparsedrive,li2024ego,jiang2023vad}. Existing methods can be broadly categorized 
based on the form of action outputs, including regression-based methods~\cite{hu2022st, jiang2023vad}, generation-based approaches~\cite{zheng2024genad,liao2025diffusiondrive}, and reinforcement-based methods~\cite{gao2025rad,yan2025adr1}. In regression-based approaches, UniAD~\cite{hu2023planning}, VAD~\cite{jiang2023vad}, and SparseDrive~\cite{sun2025sparsedrive} construct end-to-end architectures by leveraging BEV representations and vectorized or sparse queries. For generative methods, GenAD~\cite{zheng2024genad} leverages generative models for trajectory regression. DiffusionDrive~\cite{liao2025diffusiondrive} and GoalFlow~\cite{xing2025goalflow} utilize flow matching~\cite{lipman2022flow} and diffusion policy~\cite{chi2025diffusion} paradigms to sample multi-mode actions. For reinforcement learning-based approaches, recent studies~\cite{li2025driver1,jiao2025evadrive, li2025recogdrive} formulate autonomous driving as a sequential decision-making problem, which optimizes driving policies through reward design and interaction with the environment. In addition, some recent approaches incorporate large language models (LLMs) into vision-language-action (VLA) frameworks by injecting language context as high-level guidance for reasoning and decision-making~\cite{li2025recogdrive,chi2025impromptu,zhou2025opendrivevla}. 

However, most existing end-to-end driving methods lack the ability to reason about future scene dynamics under different driving actions. 
This limitation motivates the development of world models that simulate future world evolution to support more informed decision-making.

\noindent{\textbf{World Models for Autonomous Driving}}
In autonomous driving systems, the world model aims to predict the future evolution of scenes following various actions~\cite{guan2024world,kong20253d,wang2024drivingfuture,zheng2025world4drive,li2024law}. 
Combining with the next state prediction, existing world models can be categorized into two types: future scene prediction~\cite{wang2024drivingfuture,zhang2025epona,zhao2025drivedreamer4d,wang2025uniocc,zheng2024occworld} and feature-level world models in latent space~\cite{li2024law,li2025wote,li2024ssr,yang2025worldrft}. 
Representative works of future scene prediction focus on directly forecasting future observations or structured scene representations. 
For instance, GenAD~\cite{yang2024generalized}, DrivingFuture~\cite{wang2024drivingfuture}, and Epona~\cite{zhang2025epona} predict future pixel-level scenes conditioned on actions, while DriveDreamer4D~\cite{zhao2025drivedreamer4d} extends this paradigm to 4D dynamic scene generation with 4D Gaussian Splatting reconstruction. 
Other approaches like UniOcc~\cite{wang2025uniocc} and OccWorld~\cite{zheng2024occworld} operate in voxelized or occupancy space, predicting future 3D occupancy or semantic fields to model scene evolution.
Complementary to explicit scene forecasting, another line of research efforts learns the world model in a compact latent space, where the model predicts action-conditioned transitions over intermediate features rather than reconstructing full observations. 
LAW~\cite{li2024law} and World4Drive~\cite{zheng2025world4drive} construct the world model in latent feature space for planning-oriented prediction, while SSR~\cite{li2024ssr} and WoTE~\cite{li2025wote} further explore latent modeling in BEV space, moving beyond perspective-view representations.

For our design, DynFlowDrive avoids explicit pixel-level scene generation and instead operates in latent space, focusing on modeling the underlying dynamics of world evolution for action-conditioned prediction.

\begin{figure}[!ht]
    \centering
    \includegraphics[width=\textwidth]{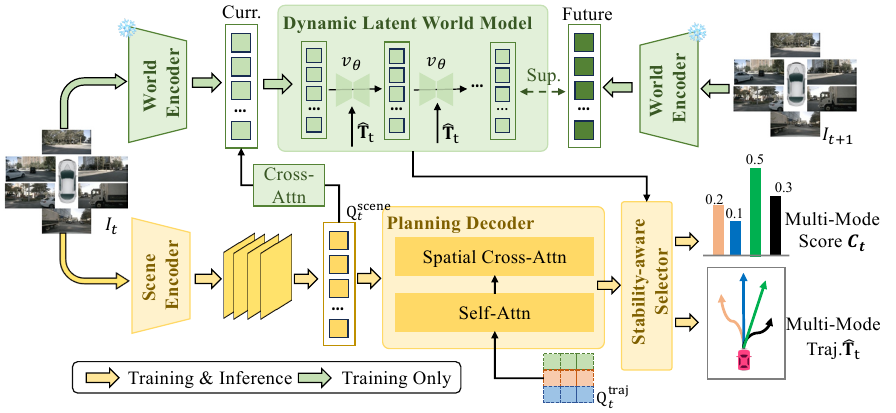}
    \vspace{-5 mm}
    \caption{\textbf{Overview of DynFlowDrive.} Given current observations, multi-mode trajectories are firstly generated by the standard planning module. A flow-based dynamic latent world model is incorporated to simulate the progressive future evolution in latent space. The resulting dynamics are used by a stability-aware multi-mode selection module, which assess the trajectory based on reconstruction quality and flow-based stability, enabling reliable supervision and improved planning robustness.}
    \vspace{-3mm}
    \label{fig:overview} %
\end{figure}

\section{Method}

As illustrated in Figure~\ref{fig:overview}, DynFlowDrive is built upon a standard end-to-end planning framework together with a flow-based latent world modeling. Given the surrounding $N$-view camera images 
$\mathbf{I}_t = \{ I_t^i \}_{i=1}^N$ and a high-level navigation command $cmd$ at current timestep $t$, a vision-based approach aims to predict the planning trajectory 
$\mathbf{T}$ that consists of a set of future points in BEV space. 

\noindent DynFlowDrive comprises three key parts:
\textbf{1) Trajectory Planning Paradigm} (Sec.~\ref{origin}), which produces multi-mode trajectories based on the observations following the standard planning paradigm in end-to-end autonomous driving;
\textbf{2) Dynamic World Model Prediction} (Sec.~\ref{dy_world}), that simulates latent scenes' evolution conditioned on the planning trajectory with the flow-based dynamic model, and 
\textbf{3) Stability-aware Multi-mode Selection} (Sec.~\ref{mode_select}), which selects trajectory modes with respect to the dynamic stability for reliable supervision. These modules work in a complementary manner, coupling dynamic world simulation with multi-mode trajectory selection.

\subsection{Trajectory Planning Paradigm}
\label{origin}
As shown in the yellow part of Figure~\ref{fig:overview}, the trajectory planning paradigm predicts multiple candidate trajectories. It first encodes the sensor inputs into a unified scene representation, 
a trajectory decoder then attends to the scene queries to iteratively refine the trajectory queries into final trajectory proposals.

\noindent{\textbf{Scene Encoder.}} Given multi-view images $\mathbf{I}_t=\{I_t^i\}_{i=1}^V$ at the current timestep $t$, each view is encoded into perspective-view features $\mathbf{F}_t^{\text{pv}} = \{\mathbf{F}_t^i\}_{i=1}^V$. A set of learnable scene queries $\mathbf{Q}^s_t$ is then introduced to aggregate multi-view information via cross-attention, which is formulated as:
\begin{equation}
\mathbf{Q}_t^{\text{scene}} = \mathrm{CrossAttn}(\mathbf{Q}_t^{\text{scene}}, \mathbf{F}_t^{\text{i}}).
\end{equation}
It yields a unified scene representation that captures the spatial layout and context from the driving scenes of the current time step.

\noindent{\textbf{Multi-modal Trajectory Decoder.}} Since $\mathbf{Q}_t^{\text{scene}}$ contains relevant scene information, following~\cite{jiang2023vad, hu2023planning}, we initialize a set of candidate waypoint queries $\mathbf{Q}_t^{\text{traj}}=\{q_t^i\}_{n=1}^{N} \in \mathbb{R}^{N_m\times N \times \times D}$ to extract multi-mode planning trajectories, where $N_m$ denotes the number of driving commands, $N$ denotes the number of trajectory modes, and $D$ is the dimensions of query feature. 

Then, scene-level context $\mathbf{Q}_t^{\text{scene}}$ is incorporated into the trajectory queries through cross-attention with the scene queries:
\begin{equation}
\mathbf{Q}_t^{\text{traj}} = \mathrm{CrossAttn}(\mathbf{Q}_t^{\text{traj}}, \mathbf{Q}_t^{\text{scene}}).
\end{equation}

Following the context-enhanced trajetcory queries, two diiferent MLP heads are applied for trajectory refinement and mode scoring. The trajectory head $\mathrm{MLP}_{\text{traj}}$ predicts a residual update to refine the waypoints, while the score head $\mathrm{MLP}_{\text{s}}$ estimates a confidence score for each mode, which can be formulated as:
\begin{equation}
\hat{\mathbf{T}}_t = \mathbf{T}_t + \mathrm{MLP}_{\text{traj}}(\mathbf{Q}_t^{\text{traj}}), \quad
\mathbf{c}_t = \mathrm{MLP}_{\text{s}}(\mathbf{Q}_t^{\text{traj}}),
\end{equation}
where $\hat{\mathbf{T}}_t=\{\hat{T}_t^n\}_{n=1}^{N}$ denotes the refined trajectories, and $\mathbf{c}_t=\{c_t^n\}_{n=1}^{N}$ represents their corresponding mode scores. These $(\hat{\mathbf{T}}_t,\mathbf{c}_t)$ pairs are treated as the multi-modal trajectory proposals for subsequent selection and training supervision.


This formulation provides a set of diverse trajectory hypotheses while maintaining a unified query-based decoding structure, which serves as the foundation for subsequent trajectory selection and dynamic world modeling.

\subsection{Dynamic Latent World Model}
\label{dy_world}

To explicitly characterize planning quality through trajectory-induced world evolution, we introduce a dynamic latent world model that simulates how the environment evolves under different trajectory hypotheses. Given the current latent state and candidate trajectories, the model predicts future latent states via, which are used to evaluate the resulting trajectory outputs in latent space.

\noindent{\textbf{World Feature Extraction.}}
For the dynamic world simulator, directly reusing the driving encoder may lead to representation shifts and unstable features for modeling temporal evolution.  Rather than follow prior works~\cite{li2024law,li2024ssr}, we employ a pretrained foundation encoder to obtain more stable and dynamics-aligned latent representations.

Given the surrounding $V$-view images $\mathbf{I}_t = \{ I_t^i \}_{i=1}^V$, we firstly encode each view into a latent representation using a pretrained VAE encoder $\mathbf{E}_{\text{vae}}$ followed by the lightweight MLP :
\begin{equation}
\mathbf{z}_t^i = \mathrm{MLP}(\mathbf{E}_{\text{vae}}(I_t^i)), \quad i = 1, \dots, V.
\end{equation}
where $\tilde{\mathbf{z}}_t^i$ denotes the dynamics-aware latent feature for the $i$-th view. Such representation benefits from the diverse training data and strong generalization capability of foundation models to obtain dynamics-oriented features.

To incorporate global scene contexts, we aggregate the multi-view latent features by interacting them with the scene queries $\mathbf{Q}_t^{\text{scene}}$ via cross-attention:
\small{
\begin{equation}
\tilde{\mathbf{z}}_t^{\text{w}} = \mathrm{CrossAttn}(\mathbf{Q}_t^{\text{scene}}, \{\mathbf{z}_t^i\}_{i=1}^V),
\end{equation}}
where $\tilde{\mathbf{z}}_t^{\text{w}}$ denotes the world latent that captures both multi-view spatial features and global scene context.

Following the same procedure, we extract the corresponding world features for the next timestep, denoted as $\tilde{\mathbf{z}}^{\text{w}}_{t+1}$, which serve as the supervision target for training the dynamic flow model.

\begin{wrapfigure}{r}{0.66\textwidth}
    \centering
    \vspace{-24pt}
\includegraphics[width=0.66\textwidth]{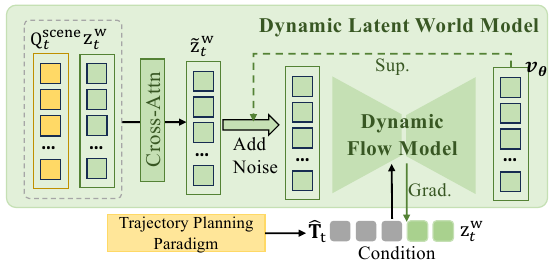}
    \vspace{-16pt}
    \caption{The architecture of our \textbf{dynamic latent world model design}, in which the velocity field $v_{\theta}$ is learnt to capture the trajectory-conditioned dynamics transitions in the latent space. }
    \vspace{-20pt}
    \label{fig:wm-design}
\end{wrapfigure}

\noindent{\textbf{Flow-based Latent World Model Simulation.}}
Given the current world latent $\tilde{\mathbf{z}}^{\text{w}}_t$ and predicted trajectories $\hat{\mathbf{T}}_t$, we model trajectory-conditioned world evolution in latent space. As shown in Figure~\ref{fig:wm-design}, instead of learning a generic mapping between noise and data, we aim to learn a trajectory-conditioned velocity field that describes how the latent state evolves under different motion hypotheses.
Following the rectified flow formulation~\cite{liu2022flow},
 we construct the noised anchor state $\mathbf{a}$ from the current latent to provide a stochastic starting point for modeling local state transitions:
 \begin{equation}
\mathbf{a} = (1-\alpha)\tilde{\mathbf{z}}_t^{\text{w}} + \alpha \boldsymbol{\epsilon}, \quad 
\boldsymbol{\epsilon} \sim \mathcal{N}(\mathbf{0}, \mathbf{I}), 
\end{equation}
where $\boldsymbol{\epsilon}$ is the Gaussian noise with the same shape as $\mathbf{z}_t^{\text{w}}$, and $\alpha \in [0,1]$ controls the perturbation strength.
Then the continuous interpolation between the anchor $\mathbf{a}$ and the target latent $\tilde{\mathbf{z}}_{t+1}^{\text{w}}$ is then defined as:
\begin{equation}
\mathbf{x}_s = (1-s)\mathbf{a} + s\,\tilde{\mathbf{z}}_{t+1}^{\text{w}}, \quad s \sim \mathcal{U}(0,1),
\end{equation}
where $s$ denotes the flow timestep.

To incorporate the trajectory conditions, planning modes $\{\hat{T}_t^n\}_{i=n}^N$ are encoded into a trajectory embeddings by $\mathrm{TrajEmb}(\cdot)$ and fused with the original VAE latent to obtain the condition features $\{\mathbf{h}_t^n\}_{n=1}^{N}$:
\begin{equation}
\mathbf{h}_t^n = \mathrm{Concat}\big([\lambda_z\cdot \mathbf{z}^{\text{w}}_t,\ \lambda_T \cdot\mathrm{TrajEmb}(\hat{T}_t^n)]\big),
\end{equation}
where $\lambda_z$ and $\lambda_T$ are scalar weights to balance the latent state and trajectory embeddings.
Then, the velocity field can be predicted by a transformer-based flow model $\mathcal{F}_\theta(\cdot)$, which is formulated as:
\begin{equation}
v_\theta = \mathcal{F}_\theta(\mathbf{x}_s, s, \mathbf{h}_t^i).
\end{equation}
This formulation models how the latent state evolves under the corresponding trajectory hypothesis.

Thus, during training, we adopt the flow matching objective that supervises the velocity along the interpolation path between the anchor state $\mathbf{a}$ and the target latent $\tilde{\mathbf{z}}_{t+1}^{\text{w}}$:
\begin{equation}
\mathcal{L}_{flow} =
\mathbb{E}
\left[
\left\|
\mathcal{F}_\theta(\mathbf{x}_s, s, \mathbf{h}_t^{i})
-
(1-s)(\tilde{\mathbf{z}}_{t+1}^{\text{w}}-\mathbf{a})
\right\|_2^2
\right],
\label{eq_loss}
\end{equation}
where the target velocity corresponds to the displacement from the anchor state toward the future latent along the interpolation path.

\noindent For sampling integration, the learned velocity defines a trajectory-conditioned dynamical system so that the future world latent can be obtained via integration:
\begin{equation}
\frac{d\tilde{\mathbf{z}}^{\text{w}}}{ds}
=
\mathcal{F}_\theta(\tilde{\mathbf{z}}^{\text{w}}(s), s, \mathbf{h}_t^{i}).
\end{equation}
Each step corresponds to an integration step along the flow variable $s$, which simulates a small portion of the latent evolution between the driving timesteps $t$ and $t+1$.

Thus, starting from the current latent $\tilde{\mathbf{z}}_t^{\text{w}}$, the future world latent can be obtained by integrating the velocity field from $s=0$ to $1$. 
In practice, we approximate this integration using a discrete Euler solver:
\begin{equation}
\tilde{\mathbf{z}}_{s_{k+1}}^{\text{w}}
=
\tilde{\mathbf{z}}_{s_k}^{\text{w}}
+
\Delta s \,
\mathcal{F}_\theta(\tilde{\mathbf{z}}_{s_k}^{\text{w}}, s_k, \mathbf{h}_t^{i}),
\end{equation}
where $s_k$ is the corresponding flow timestep, and $\Delta s$ is the integration step size.

This formulation models how different trajectory hypotheses induce distinct latent state transitions, enabling dynamics-aware trajectory evaluation based on the progressive evolution dynamics.

\begin{wrapfigure}{r}{0.35\textwidth}
    \centering
    \vspace{-56pt}
\includegraphics[width=0.35\textwidth]{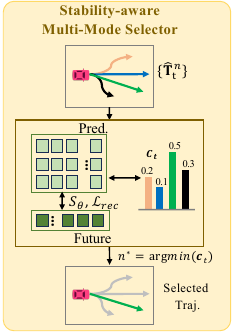}
    \vspace{-16pt}
    \caption{Stability-aware Multi-mode Selection. For training, the score head is supervised by the stable criterion. For Inference, the best mode trajectory is selected according to the highest score index.} 
    \vspace{-26pt}
    \label{fig:selector}
\end{wrapfigure}

\subsection{Stability-aware Multi-mode Selection}
\label{mode_select}

Our multi-modal trajectory predictor generates a set of candidate trajectories 
$\{\hat{\mathbf{T}}_t^{n}\}_{n=1}^{N}$ to capture the inherent uncertainty of future driving behaviors. 
Conventional selection strategies typically rely on geometric or reconstruction criteria, 
such as trajectory error (e.g., ADE/FDE) and latent reconstruction loss. 
However, these metrics only measure spatial or feature-level similarity and fail to reflect the quality of trajectory-induced world evolution.

In practice, different trajectory candidates may produce similar geometric errors while leading to substantially different latent state evolutions. 
Selecting trajectories solely based on geometric criteria may therefore favor modes that are spatially accurate but induce unstable or unrealistic scene dynamics, which is undesirable for reliable planning.

To address this issue, as shown in Figure~\ref{fig:selector}, we introduce a stability-aware selection strategy that evaluates each trajectory based on the induced dynamics.
Given a trajectory candidate $\hat{\mathbf{T}}_t^{n}$, the flow-based world model simulates the latent evolution conditioned on this trajectory and produces a sequence of velocities 
$\{\mathbf{v}_k^{n}\}_{k=1}^{K}$ along the integration steps, where $K$ is the discrete steps during transitions. 
We measure the directional consistency of the evolution by computing the average angular deviation between consecutive normalized velocity directions $\hat{\mathbf{u}}_{k-1}^{n}$ and $\hat{\mathbf{u}}_{k}^{n}$:
\begin{equation}
\hat{\mathbf{u}}_k^{n} = \frac{\mathbf{v}_k^{n}}{\|\mathbf{v}_k^{n}\|}, \quad
\mathcal{S}_{\theta}^{n} = \frac{1}{K-1}\sum_{k=2}^{K}\arccos\!\left(\hat{\mathbf{u}}_k^{n} \cdot \hat{\mathbf{u}}_{k-1}^{n}\right),
\end{equation}
where smaller $\mathcal{S}_{\theta}^{n}$ value indicates smoother and more physically consistent dynamics.

Based on this, we define the unified selection criterion, combined with other commonly used metrics~\cite {zheng2025world4drive} for each trajectory mode assessment:
\begin{equation}
\mathcal{C}^{n} = 
\lambda_{rec} \, \mathcal{L}_{rec}^{n} 
+ \lambda_{traj} \, \mathcal{L}_{traj}^{n} 
+ \lambda_{\theta} \, \mathcal{S}_{\theta}^{n}, 
\quad
n^* = \arg\min_n \mathcal{C}^{n},
\end{equation}
where $\mathcal{L}_{traj}^{n}$ denotes the trajectory error between the predicted trajectory and the ground truth, and $\mathcal{L}_{rec}^{n}$ measures the reconstruction discrepancy of the predicted latent state.
The selected optimal mode $n^*$ is employed to supervise the learning of the trajectory scoring head.

By explicitly incorporating trajectory-conditioned dynamical stability, the proposed strategy favors trajectories that induce smooth and physically consistent world evolution, leading to more reliable trajectory assessment beyond conventional criteria.

\subsection{Training and Inference Settings}

For model training, we jointly optimize trajectory prediction, scoring, latent reconstruction, and flow-based dynamics with a unified objective:
\begin{equation}
\mathcal{L} = 
\mathcal{L}_{traj} 
+ \lambda_{score}\mathcal{L}_{score} 
+ \lambda_{rec}\mathcal{L}_{rec} 
+ \lambda_{flow}\mathcal{L}_{flow}.
\end{equation}

Here, $\mathcal{L}_{traj}$ denotes the $L_1$ loss between predicted trajectories and the ground-truth future trajectory. 
The scoring loss $\mathcal{L}_{score}$ is supervised using the selected optimal mode $n^*$ defined in Sec.~\ref{mode_select}. 
$\mathcal{L}_{rec}$ enforces consistency between the predicted latent state and the target latent representation, while $\mathcal{L}_{flow}$ corresponds to the flow matching objective used to learn the trajectory-conditioned velocity field. 
The coefficients $\lambda_{score}$, $\lambda_{rec}$, and $\lambda_{flow}$ balance the contributions of different objectives.

During inference, the world model is not involved, since the driving system has already acquired the capability to select the optimal trajectory. 
The planning module directly selects the trajectory with the highest predicted score as the final output, introducing no additional computational overhead compared with the standard trajectory prediction framework.

\section{Experiment}

\subsection{Dataset and Metrics}
For comprehensive evaluations, we train and evaluate our DynFlowDrive on two benchmarks, including the open-loop nuScenes~\cite{caesar2020nuscenes} and the closed-loop NavSim~\cite{dauner2024navsim}.

\noindent{\textbf{Open-Loop nuScenes Benchmark.}} The nuScenes dataset contains 1000 driving scenes across different driving scenarios in Singapore and Boston. In line with previous studies~\cite{li2024law, li2024ssr, li2024ego}, we make use of the calculated displacement error ($L_2$) with the ground truth and collision rate (CR) for performance evaluation. Displacement error is calculated by $L_2$ error between the prediction and the GT trajectory, while CR quantifies the collision probability with surrounding objects following the predicted trajectories. All metrics are calculated within the 3s future with a 0.5s interval and evaluated at 1s, 2s, and 3s. We utilize the VAD evaluation metrics~\cite{jiang2023vad} that compute the average across all previous frames.

\noindent{\textbf{Closed-loop NavSim Benchmark.}} NavSim provides a closed-loop benchmark built on the OpenScene dataset. It contains 103K keyframes with 1192 training scenarios and 136 test scenarios. During evaluation, the predicted trajectories are interpolated using an LQR controller sampled at 2 Hz over a 4-second horizon. It is evaluated by a simulator to get the rule-based simulation metric score, which is calculated based on five key metrics, including No at-fault Collisions (NC),
Drivable Area Compliance (DAC), Time to Collision with
bounds (TTC), Ego Progress (EP), and Comfort (C). Thus, the final
PDM Score (PDMS) is derived by aggregating these metrics as follows
$
    \text{PDMS} = \text{NC} \times \text{DAC} \times \frac{5 \times (\text{EP} + \text{TTC}) + 2 \times \text{C}}{12}
$.
\subsection{Implementation Details}

\noindent{\textbf{nuScenes.}} For open-loop evaluation, our DynFlowDrive is trained for 12 epochs on 8 NVIDIA RTX 4090 GPUs with a batch size of 1 per GPU, which is built on the open-sourced LAW~\cite{li2024law}, SSR~\cite{li2024ssr}. The driving commands align with previous works~\cite{hu2023planning}, including left, right, and go straight. For the SSR-based approaches, we adopt
ResNet50~\cite{he2016deep} as image backbone operating at an image resolution of 640 × 360. We utilize the AdamW optimizer~\cite{loshchilov2017decoupled} with a learning rate set to $1 \times 10^{-4}$. The weights $\lambda_{score}$ for scoring loss, $\lambda_{rec}$ for latent reconstruction loss, and $\lambda_{flow}$ for flow loss are set to 0.5, 0.2, and 0.1, respectively.  

\noindent{\textbf{NavSim.}} We further conduct the closed loop evaluation of DynFlowDrive on the NavSim benchmark. Our model is trained for 30 epochs with a batch size of 16 per GPU. Following the setting of WoTE~\cite{li2025wote}, we concatenate the front-view image with center-cropped
front-left and front-right images, resulting in a combined resolution of 256 × 1024 pixels. For LiDAR, the 64m × 64m point cloud surrounding the ego vehicle is used. For network architecture, we employ ResNet34~\cite{he2016deep} as the backbone for image feature extraction. The number of trajectory anchors is set to $N=256$. We utilize the Adam optimizer with a learning rate of $1 \times 10^{-4}$.

\begin{table}[!t]
    \begin{center}
    \setlength{\tabcolsep}{6pt} 
    \renewcommand{\arraystretch}{1.1}
    \resizebox{\linewidth}{!}{
    \begin{tabular}{l|c|c|cccc|cccc}
        \toprule
        \multirow{2}{*}{\textbf{Method}} & \multirow{2}{*}{\textbf{Pub.}} & \multirow{2}{*}{\textbf{BEV}} & \multicolumn{4}{c}{\textbf{$\mathbf{L_2}$ (m) $\downarrow$}} & \multicolumn{4}{c}{\textbf{Collision Rate (\%) $\downarrow$}}  \\
        \cmidrule(lr){4-7} \cmidrule(lr){8-11}
        & & & \textbf{1s} & \textbf{2s} & \textbf{3s} & \textbf{Avg.} & \textbf{1s} & \textbf{2s} & \textbf{3s} & \textbf{Avg.}  \\
        \midrule
       
        \multicolumn{11}{c}{\textit{Perception-based Approaches}}\\
        \midrule
        ST-P3~\cite{hu2022st} & ECCV 2022 & \ding{51} & 1.33 & 2.11 & 2.90 & 2.11 & 0.23 & 0.62 & 1.27 & 0.71  \\
        UniAD~\cite{hu2023planning} & CVPR 2023  &\ding{51}  & 0.48 & 0.96 & 1.65 & 1.03 & 0.05 & 0.17 & 0.71 & 0.31  \\
        VAD-Tiny~\cite{jiang2023vad} & ICCV 2023 & \ding{51} & 0.46 & 0.76 & 1.12 & 0.78 & 0.21 & 0.35 & 0.58 & 0.38  \\
        VAD-Base~\cite{jiang2023vad} & ICCV 2023 &\ding{51}  & 0.41 & 0.70 & 1.05 & 0.72 & 0.07 & 0.17 & 0.41 & 0.22  \\
        BEV-Planner~\cite{li2024ego} & CVPR 2024 &\ding{51}  & 0.28 & 0.42 & 0.68 & 0.46 & 0.04 & 0.37 & 1.07 & 0.49  \\
        PARA-Drive~\cite{weng2024paradrive} & CVPR 2024 & \ding{51} & 0.25 & 0.46 & 0.74 & 0.48 & 0.14 & 0.23 & 0.39 & 0.25  \\
        GenAD~\cite{zheng2024genad} & ECCV 2024 &\ding{51}  & 0.28 & 0.49 & 0.78 & 0.52 & 0.08 & 0.14 & 0.34 & 0.19  \\
        SparseDrive~\cite{sun2025sparsedrive} & ICCV 2025 & \ding{55} & 0.29 & 0.58 & 0.96 & 0.61 & 0.01 & 0.05 & 0.18 & 0.08  \\
        Drive-OccWorld~\cite{yang2025drivingoccworld} & AAAI 2025 & \ding{55}& 0.25 & 0.44 & 0.72 & 0.47 & 0.03 & 0.08 & 0.22 & 0.11  \\
        MomAD~\cite{song2025momoad} & CVPR 2025 &\ding{55} & 0.31 & 0.57 & 0.91 & 0.60 & 0.01 & 0.05 & 0.22 & 0.09 \\
        DiffusionDrive~\cite{liao2025diffusiondrive} & CVPR 2025 &\ding{55} & 0.27 & 0.54 & 0.90 & 0.57 & 0.03 & 0.05 & 0.16 & 0.08\\
        \midrule
       
        \multicolumn{11}{c}{\textit{Latent World Model Approaches}}\\
        \midrule
        LAW~\cite{li2024law} & ICLR 2025 &\ding{55} & 0.26 & 0.57 & 1.01 & 0.61 & 0.14 & 0.21 & 0.54 & 0.30  \\
        \textbf{DynFlowDrive}(LAW) & - & \ding{55}& \textbf{0.24} & \textbf{0.54} & \textbf{0.92} & \textbf{0.57} & \textbf{0.11} & \textbf{0.12} & \textbf{0.48} & \textbf{0.22} \\
        \midrule
        
        SSR$^\ast$~\cite{li2024ssr} & ICLR 2025 &\ding{51} & 0.18 & 0.35 & 0.63 & 0.39 & 0.08 & 0.12 & 0.24 & 0.15 \\
        \textbf{DynFlowDrive} (SSR) & - &\ding{51} & 0.16 & 0.32 & 0.57 & 0.35 & 0.07 & 0.09 & 0.21 & 0.14 \\
        \textbf{DynFlowDrive}$\diamond$ (SSR) & - &\ding{51} & \textbf{0.13} & \textbf{0.27} & \textbf{0.52} & \textbf{0.31} & \textbf{0.06} & \textbf{0.09} & \textbf{0.18} & \textbf{0.11} \\
        \bottomrule
    \end{tabular}
    }
     \end{center}
     \vspace{-2mm}
    \caption{\textbf{Comparison of the SOTA methods on the nuScenes dataset.} ``$\ast$'' denotes the re-implemented result from SSR~\cite{li2024ssr} by official public codes. ``$\diamond$'' denotes using ego status in the planning module following BEVPlanner++~\cite{li2024ego}. Evaluations are conducted in the same way as the settings in VAD~\cite{jiang2023vad}. Our DynFLowDrive are implemented based on LAW~\cite{li2024law} and SSR. LAW adopts Swin-Tiny~\cite{liu2021swin} as image backbone, while others adopt ResNet-50~\cite{he2016deep}.
    }
    \label{tab:nuscenes}
   \vspace{-10mm}
\end{table}

\subsection{Main Results}

\noindent{\textbf{Open-Loop Evaluation.}} On the nuScenes~\cite{caesar2020nuscenes} benchmark, we compare our proposed DynFlowDrive with the prior approaches following the evaluation metrics in VAD~\cite{jiang2023vad}. As shown in Table~\ref{tab:nuscenes}, our method consistently achieves superior results in both the $L_2$ distance error and collision rate. We implement our DynFlowDrive based on two representative latent world model baselines. For the approaches without BEV representation, compared with LAW~\cite{li2024law}, our DynFlowDrive reduces the $L_2$ displacement error by 0.4m and lowers the collision rate by 26\%. Based on the BEV representation, we replace the latent world model in SSR~\cite{li2024ssr} with our DynFlowDrive design, which yields nearly a 20\% reduction on $L_2$ distance by 0.08m and achieves 0.11\% collision rate. It indicates that our approach can be seamlessly integrated into existing frameworks and consistently improve planning quality. Compared to the perception-based approaches, without the incorporation of any auxiliary tasks, our dynamic latent world model still achieves comparable performance.

\begin{figure}[!t]
    \centering
    \includegraphics[width=\textwidth]{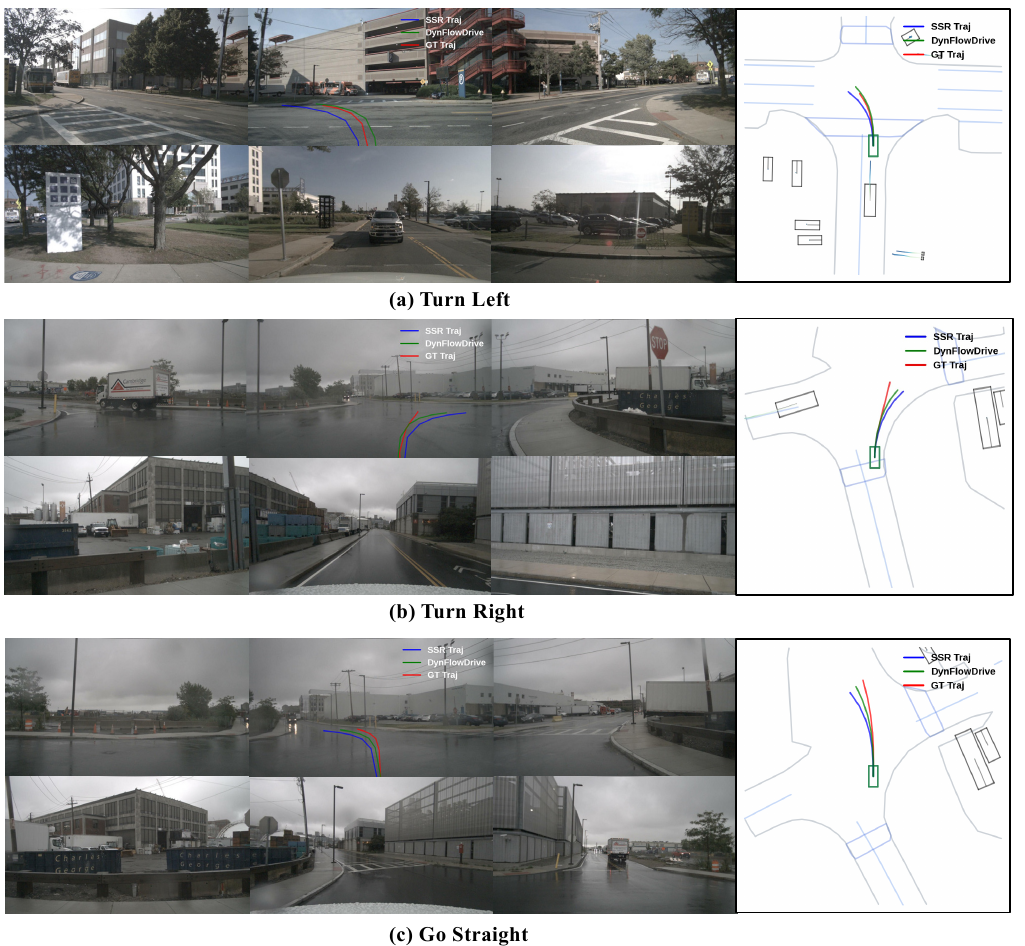}
    \vspace{-4mm}
    \caption{\textbf{Visualization of Planning Results on nuScenese dataset.} The perception representations are rendered from annotations. Surrounding light lines denote the maps of the scene, while black boxes represent the object detection. The ego vehicle is drawn by the green box in the center.}%
    \vspace{-4mm}
    \label{fig:visual} %
\end{figure}

\noindent{\textbf{Qualitative Result.}} In Figure~\ref{fig:visual}, we present the qualitative comparisons between our DynFlowDrive and the baseline SSR with three different commands. As illustrated in both the BEV space and front-view camera perspectives, our planning results exhibit stronger alignment with the ground truth of the scene structure. In particular, our method produces smoother and more coherent trajectories that better follow lane geometry and surrounding dynamics, indicating improved consistency between predicted motions and the underlying environment. More visualizations are provided in our supplementary materials.

\noindent{\textbf{Closed-Loop Evaluation.}} 
We further evaluate DynFlowDrive under closed-loop settings on the NavSim benchmark~\cite{dauner2024navsim}. 
The comparison results are summarized in Table~\ref{tab:sota_NavSim}. 
As shown, DynFlowDrive achieves competitive performance across closed-loop metrics and reaches the state-of-the-art planning accuracy of 88.7\% PDMS, surpassing the baseline world model approach WoTE~\cite{li2025wote}. 
Notably, without relying on auxiliary tasks such as object detection or online mapping, our method also outperforms previous world model-based approaches, including LAW~\cite{li2024law} and World4Drive~\cite{zheng2025world4drive}. 
These results demonstrate that incorporating the flow-based latent world model together with the stability-aware selection mechanism enables the model to better capture trajectory-conditioned world dynamics, leading to safer and more reliable planning behavior in closed-loop driving scenarios.

\begin{table*}[!t]
    \begin{center}
    \renewcommand{\arraystretch}{1.1}
        \resizebox{\linewidth}{!}{
            \begin{tabular}{l|c|c|c|ccccc|c}
                \toprule
                \textbf{Method} & \textbf{Input} & \textbf{\#Traj.} & \textbf{Traj. Eval.} & \textbf{NC} $\uparrow$ & \textbf{DAC} $\uparrow$ & \textbf{EP} $\uparrow$ & \textbf{TTC} $\uparrow$ & \textbf{Comf.} $\uparrow$ & \textbf{PDMS} $\uparrow$ \\ 
                \midrule
                Human & - & - & - & 100 & 100 & 87.5 & 100 & 99.9 & 94.8 \\ 
                \midrule
                Constant Velocity & - & 1 & \ding{55} & 69.9 & 58.8 & 49.3 & 49.3 & 100.0 & 21.6 \\ 
                
                Ego Status MLP & - & 1 & \ding{55} & 93.0 & 77.3 & 62.8 & 83.6 & 100.0 & 65.6 \\ 
                VADv2~\cite{chen2024vadv2} & C & 8192 & Rule-based & 97.9 & 91.7 & 77.6 & 92.9 & 100.0 & 83.0 \\  
                UniAD~\cite{hu2023planning} & C & 1 & Rule-based & 97.8 & 91.9 & 78.8 & 92.9 & 100.0 & 83.4 \\ 
                LTF~\cite{prakash2021multi} & C & 1 & \ding{55} & 97.4 & 92.8 & 79.0 & 92.4 & 100.0 & 83.8 \\ 
                PARA-Drive~\cite{weng2024paradrive} & C & 1 & Rule-based & 97.9 & 92.4 & 79.3 & 93.0 & 99.8 & 84.0 \\ 
                TransFuser~\cite{chitta2022transfuser} & C \& L & 1 & \ding{55} & 97.7 & 92.8 & 79.2 & 92.8 & 100.0 & 84.0 \\
                LAW~\cite{li2024law} & C & 1 & \ding{55} & 96.4 & 95.4 & 81.7 & 88.7 & 99.9 & 84.6 \\
                World4Drive~\cite{zheng2025world4drive}&C&6&\ding{55}&97.4&94.3&79.9&92.8&100.0&85.1\\
                DRAMA~\cite{yuan2024drama} & C \& L  & 1 & \ding{55} & 98.0 & 93.1 & 80.1 & 94.8 & 100.0 & 85.5 \\
                Hydra-MDP~\cite{li2024hydra} & C \& L  & 8192 & Model-free & 98.3 & 96.0 & 78.7 & 94.6 & 100.0 & 86.5 \\
                DiffusionDrive~\cite{liao2025diffusiondrive}&C \& L&20&\ding{55}&98.2&96.2&82.2&94.7&100.0&88.1\\
                WoTE~\cite{li2025wote} & C \& L  & 256 & Model-based & 98.5 & 96.8 & 81.9 & 94.9 & 99.9 & 88.3 \\ 
                \midrule
                \textbf{DynFlowDrive (Ours)}&C \& L  & 256 & Model-based  &  \textbf{98.7} & \textbf{96.8}  & \textbf{82.5}  & \textbf{95.5}  &  \textbf{100.0} &  \textbf{88.7} \\ 
                \bottomrule
            \end{tabular}
        }
    \vspace{1mm}
    \caption{\textbf{Comparison with the SOTA approaches on the NavSim test set.} 
    C: Camera.
    L: LiDAR.
    Rule-based means that trajectory evaluation follows specific rules.
    Model-free represents the evaluation without world modeling.
    Model-based denotes evaluating the trajectory with the assistance of the world model.
    }
     \label{tab:sota_NavSim}
     \vspace{-8mm}
    \end{center}
\end{table*}

\subsection{Ablation Study}
In this section, we conduct comprehensive ablation studies to investigate the effectiveness of our DynFlowDrive designs. For fair comparison, all ablations are conducted on the nuScenes benchmark. More ablations and analyses on NavSim benchmark are provided in our supplementary materials.

\noindent{\textbf{Verification of Component Designs.}}
In this section, we conduct ablation studies to evaluate the contribution of each component and the combinations. As shown in Table~\ref{Component-Ablation}, starting from the vanilla end-to-end autonomous driving paradigm without BEV representation or world modeling modules, incorporating the static transformer-based world model yields only marginal improvement. By introducing the proposed dynamic latent world model, we achieve an average $L_2$ error reduction of 0.1\,m and a 11\% decrease in collision rate, demonstrating the benefit of dynamic simulation of world evolution. Furthermore, when combined with the stability-aware multi-mode selection, DynFlowDrive achieves the best overall performance, reaching an $L_2$ error of 0.57\,m and a collision rate of 0.22\%. These results validate that dynamic evolution modeling and stability-guided mode selection jointly contribute to safer and more accurate trajectory planning. For inference, no additional computational burdens are introduced, as the flow-based world model is only used during training and does not participate in the forward prediction pipeline. 
\begin{table}[t]
    \centering
    \setlength{\tabcolsep}{8pt} 
    \resizebox{0.98\linewidth}{!}{
    \begin{tabular}{ccc cccc cccc c}
        \toprule
        \multicolumn{3}{c}{\textbf{Modules}} & \multicolumn{4}{c}{$L_2$ (m) $\downarrow$} & \multicolumn{4}{c}{CR (\%) $\downarrow$}&\multirow{2}{*}{FPS $\uparrow$} \\
        Static-WM&Dyn-WM & MS & 1s & 2s & 3s & Avg. & 1s & 2s & 3s & Avg.& \\
        \midrule
        -&-&-&0.31&0.63&1.12&0.69&0.19&0.27&0.64&0.37&13.8\\
        $\checkmark$&- &  - &  0.26 & 0.57  & 1.01  & 0.61  & 0.14  &0.21 & 0.54 & 0.30 &13.8\\ 
        -&$\checkmark$ & -  &  0.26 &  0.55 & 0.94  & 0.59  & 0.11  & 0.16 & 0.51  & 0.26& 13.8\\
        -&- & $\checkmark$ &  0.25  &  0.55 & 0.93  & 0.58  & 0.11  &0.15  & 0.49  &0.25&13.6\\ 
        -& $\checkmark$& $\checkmark$ & \textbf{0.24} &  \textbf{0.54} &  \textbf{0.92} & \textbf{0.57}  & \textbf{0.10} & \textbf{0.12} & \textbf{0.45} & \textbf{0.22}&13.6\\ 
        \bottomrule
    \end{tabular}}
    \vspace{1mm}
    \caption{\textbf{Component-wise Ablation.} Static-WM and Dyn-WM mean the existing stable world model design and our flow-based dynamic modeling. MS denotes the strategy of stability-aware multi-mode selection. FPS are measured on one RTX 4090.}
    \vspace{-8mm}
    \label{Component-Ablation}
\end{table}

\noindent{\textbf{Ablations on the Dynamic Latent World Model Design.}}
We evaluate different world model designs in Table~\ref{ablation-wm}. 
Replacing the static transformer-based world model with the proposed flow-based formulation consistently improves long-horizon prediction. 
In particular, the 3s $L_2$ error is reduced from 1.01\,m to 0.94\,m and the collision rate decreases from 0.54\% to 0.51\%. 
This improvement indicates that flow-based modeling better captures the dynamic evolution of the driving scene in latent space, resulting in smoother and more coherent trajectory predictions.
Furthermore, applying the flow-based dynamics on world features extracted by the pretrained encoder brings additional gains.
By decoupling representation learning from dynamic evolution modeling, the model achieves more stable latent transitions, reducing the average $L_2$ error to 0.57\,m and the collision rate to 0.22\%. 
These results demonstrate that combining flow-based dynamics with stable latent representations leads to more reliable trajectory-conditioned world modeling.

\begin{table}[t]
    \centering
    \setlength{\tabcolsep}{8pt} 
    \resizebox{0.95\linewidth}{!}{
    \begin{tabular}{l|cccc|cccc}
        \toprule
        \multirow{2}{*}{\textbf{World Model Design}} & \multicolumn{4}{c|}{$L_2$ (m) $\downarrow$} & \multicolumn{4}{c}{CR (\%) $\downarrow$} \\
        &  1s & 2s & 3s & Avg. & 1s & 2s & 3s & Avg. \\
        \midrule
        Static WM &  0.26 & 0.57  & 1.01  & 0.61  & 0.14  &0.21 & 0.54 & 0.30 \\ 
        Flow WM & 0.26  & 0.55  & 0.94  & 0.59  & 0.11  & 0.16  & 0.51  & 0.26 \\ 
        \quad + World Feat. Design & \textbf{0.24}  & \textbf{0.54}  & \textbf{0.92}  & \textbf{0.57}  & \textbf{0.10}  &  \textbf{0.12} & \textbf{0.45}  & \textbf{0.22} \\ 
        \bottomrule
    \end{tabular}}
    \vspace{1mm}
    \caption{\textbf{Ablations on the Dynamic Latent World Model Design.}}
    \vspace{-8mm}
    \label{ablation-wm}
\end{table}

\noindent{\textbf{Ablations on Flow Integration Steps.}}
We study the impact of the number of integration steps used to approximate the continuous flow dynamics. 
As shown in Table~\ref{ablation-diffusion}, increasing the number of steps from 1 to 5 consistently improves both trajectory accuracy and safety, reducing the $L_2$ error from 0.60\,m to 0.57\,m and the collision rate from 0.28\% to 0.22\%. 
This demonstrates that a finer discretization better captures the underlying trajectory-conditioned dynamics.
However, further increasing the number of steps to 10 yields no additional gains and even slightly degrades performance. 
We attribute this to accumulated numerical errors and over-smoothing during integration. 
Overall, a moderate number of integration steps of 5 achieves the best trade-off between accuracy and stability.

\begin{table}[!h]
\vspace{-4mm}
    \centering

    \begin{subtable}{0.48\linewidth}
        \centering
        \setlength{\tabcolsep}{6pt} 
        \resizebox{\linewidth}{!}{
        \begin{tabular}{c|cccc}
            \toprule
            \textbf{\# Steps} &1& 3 & 5 & 10  \\
            \midrule
            $L_2$ (m) Avg.$\downarrow$ & 0.60 & 0.59&  \textbf{0.57} & 0.59 \\
            CR (\%) Avg.$\downarrow$ & 0.28 & 0.23 & \textbf{0.22} & 0.24 \\
            \bottomrule
        \end{tabular}}
        \caption{Ablations on Flow Integration Steps.}
        \label{ablation-diffusion}
    \end{subtable}
    \hfill
    \begin{subtable}{0.48\linewidth}
        \centering
        \setlength{\tabcolsep}{6pt} 
        \resizebox{\linewidth}{!}{
        \begin{tabular}{c|cc}
            \toprule
            \textbf{Selection Strategy} &  $L_2$ (m) Avg. $\downarrow$ & CR (\%) Avg. $\downarrow$ \\
            \midrule
            / & 0.61 & 0.30 \\
            $L_2$ Only & 0.59 & 0.24 \\
            + Recons. & 0.58 & 0.22 \\
            + Flow Stab. & \textbf{0.57} & \textbf{0.22} \\
            \bottomrule
        \end{tabular}}
        \caption{Ablations on Mode Selection Criteria.}
        \label{ablation-select}
    \end{subtable}

    \vspace{-2mm}
    \caption{\textbf{Ablations on Flow Steps and Multi-Mode Selection Strategies.}}
    \vspace{-8mm}
    \label{ablation-diffusion-select}
\end{table}

\noindent{\textbf{Ablations on Multi-Mode Selection.}}
We study different multi-mode selection strategies in Table~\ref{ablation-select}. Selecting modes solely based on $L_2$ distance is suboptimal, as it emphasizes point-wise accuracy while neglecting temporal consistency and physical plausibility, leading to higher collision rates. 
Incorporating reconstruction similarity improves latent alignment and trajectory coherence, yielding consistent gains. Further introducing flow direction as a stability-aware criterion encourages modes to follow the learned dynamic evolution, resulting in the lowest collision rate of 0.22\% with competitive $L_2$ distance of 0.57 m. 
These results show that combined with the reconstruction similarity, flow-based stability provides a more reliable selection criterion for safe and robust trajectory planning.

\section{Conclusion}
In this work, we propose \textbf{DynFlowDrive}, a flow-based dynamic latent world modeling that explicitly models the progressive scene evolution in latent space via rectified flow. By conditioning on candidate trajectories, our approach enables dynamic simulation of future world states and supports more reliable trajectory evaluation through stability-aware multi-mode selection.
Extensive experiments on nuScenes and NavSim demonstrate that DynFlowDrive consistently improves planning accuracy and safety across different end-to-end driving frameworks. These results highlight the importance of explicitly modeling world dynamics for robust and scalable autonomous driving systems. 
Future work includes integrating vision-language models (VLMs) for stronger semantic reasoning, improving robustness under rare corner cases, and ensuring scalability to diverse driving environments.

\clearpage  %

\bibliographystyle{splncs04}
\bibliography{main}

@String(PAMI  = {IEEE Trans. Pattern Anal. Mach. Intell.})

@String(CVPR  = {IEEE Conf. Comput. Vis. Pattern Recog.})

@String(ICCV  = {Int. Conf. Comput. Vis.})

@String(ECCV  = {Eur. Conf. Comput. Vis.})

@String(NeurIPS = {Adv. Neural Inform. Process. Syst.})

@String(ICLR  = {Int. Conf. Learn. Represent.})

@String(AAAI  = {AAAI})

@String(IJCAI = {IJCAI})

@string(TIV = "IEEE Trans. Intell. Veh.")

@string{ICRA = "IEEE Int. Conf. on Rob. and Auto."}

@String(PAMI  = {IEEE TPAMI})

@String(CVPR  = {CVPR})

@String(ICCV  = {ICCV})

@String(ECCV  = {ECCV})

@String(NeurIPS = {NeurIPS})

@String(ICLR  = {ICLR})

@string(ICRA = {ICRA})

@String(TIV = {TIV})

@article{li2024bevformer,
  title={Bevformer: learning bird's-eye-view representation from lidar-camera via spatiotemporal transformers},
  author={Li, Zhiqi and Wang, Wenhai and Li, Hongyang and Xie, Enze and Sima, Chonghao and Lu, Tong and Yu, Qiao and Dai, Jifeng},
  journal=PAMI,
  year={2024},
  publisher={IEEE}
}

@inproceedings{liu2024mgmap,
  title={Mgmap: Mask-guided learning for online vectorized hd map construction},
  author={Liu, Xiaolu and Wang, Song and Li, Wentong and Yang, Ruizi and Chen, Junbo and Zhu, Jianke},
  booktitle=CVPR,
  pages={14812--14821},
  year={2024}
}

@inproceedings{liu2025uncertainty,
  title={Uncertainty-Instructed Structure Injection for Generalizable HD Map Construction},
  author={Liu, Xiaolu and Yang, Ruizi and Wang, Song and Li, Wentong and Chen, Junbo and Zhu, Jianke},
  booktitle=CVPR,
  pages={22359--22368},
  year={2025}
}

@inproceedings{wang2024not,
  title={Not all voxels are equal: Hardness-aware semantic scene completion with self-distillation},
  author={Wang, Song and Yu, Jiawei and Li, Wentong and Liu, Wenyu and Liu, Xiaolu and Chen, Junbo and Zhu, Jianke},
  booktitle=CVPR,
  pages={14792--14801},
  year={2024}
}

@inproceedings{hu2023planning,
  title={Planning-oriented autonomous driving},
  author={Hu, Yihan and Yang, Jiazhi and Chen, Li and Li, Keyu and Sima, Chonghao and Zhu, Xizhou and Chai, Siqi and Du, Senyao and Lin, Tianwei and Wang, Wenhai and others},
  booktitle=CVPR,
  pages={17853--17862},
  year={2023}
}

@inproceedings{sun2025sparsedrive,
  title={Sparsedrive: End-to-end autonomous driving via sparse scene representation},
  author={Sun, Wenchao and Lin, Xuewu and Shi, Yining and Zhang, Chuang and Wu, Haoran and Zheng, Sifa},
  booktitle=ICRA,
  pages={8795--8801},
  year={2025},
  organization={IEEE}
}

@inproceedings{li2024ego,
  title={Is ego status all you need for open-loop end-to-end autonomous driving?},
  author={Li, Zhiqi and Yu, Zhiding and Lan, Shiyi and Li, Jiahan and Kautz, Jan and Lu, Tong and Alvarez, Jose M},
  booktitle=CVPR,
  year={2024}
}

@inproceedings{jiang2023vad,
  title={Vad: Vectorized scene representation for efficient autonomous driving},
  author={Jiang, Bo and Chen, Shaoyu and Xu, Qing and Liao, Bencheng and Chen, Jiajie and Zhou, Helong and Zhang, Qian and Liu, Wenyu and Huang, Chang and Wang, Xinggang},
  booktitle=ICCV,
  pages={8340--8350},
  year={2023}
}

@article{gao2025rad,
  title={RAD: Training an End-to-End Driving Policy via Large-Scale 3DGS-based Reinforcement Learning},
  author={Gao, Hao and Chen, Shaoyu and Jiang, Bo and Liao, Bencheng and Shi, Yiang and Guo, Xiaoyang and Pu, Yuechuan and Yin, Haoran and Li, Xiangyu and Zhang, Xinbang and Zhang, Ying and Liu, Wenyu and Zhang, Qian and Wang, Xinggang},
  journal={arXiv preprint arXiv:2502.13144},
  year={2025}
}

@inproceedings{zheng2024genad,
  title={Genad: Generative end-to-end autonomous driving},
  author={Zheng, Wenzhao and Song, Ruiqi and Guo, Xianda and Zhang, Chenming and Chen, Long},
  booktitle=ECCV,
  pages={87--104},
  year={2024},
  organization={Springer}
}

@inproceedings{yang2024generalized,
  title={Generalized predictive model for autonomous driving},
  author={Yang, Jiazhi and Gao, Shenyuan and Qiu, Yihang and Chen, Li and Li, Tianyu and Dai, Bo and Chitta, Kashyap and Wu, Penghao and Zeng, Jia and Luo, Ping and others},
  booktitle={CVPR},
  pages={14662--14672},
  year={2024}
}

@inproceedings{liao2025diffusiondrive,
  title={Diffusiondrive: Truncated diffusion model for end-to-end autonomous driving},
  author={Liao, Bencheng and Chen, Shaoyu and Yin, Haoran and Jiang, Bo and Wang, Cheng and Yan, Sixu and Zhang, Xinbang and Li, Xiangyu and Zhang, Ying and Zhang, Qian and others},
  booktitle=CVPR,
  pages={12037--12047},
  year={2025}
}

@inproceedings{hu2022st,
  title={St-p3: End-to-end vision-based autonomous driving via spatial-temporal feature learning},
  author={Hu, Shengchao and Chen, Li and Wu, Penghao and Li, Hongyang and Yan, Junchi and Tao, Dacheng},
  booktitle=ECCV,
  pages={533--549},
  year={2022},
  organization={Springer}
}

@article{yan2025adr1,
  title={AD-R1: Closed-loop reinforcement learning for end-to-end autonomous driving with impartial world models},
  author={Yan, Tianyi and Tang, Tao and Gui, Xingtai and Li, Yongkang and Zhesng, Jiasen and Huang, Weiyao and Kong, Lingdong and Han, Wencheng and Zhou, Xia and Zhang, Xueyang and others},
  journal={arXiv preprint arXiv:2511.20325},
  year={2025}
}

@inproceedings{xing2025goalflow,
  title={Goalflow: Goal-driven flow matching for multimodal trajectories generation in end-to-end autonomous driving},
  author={Xing, Zebin and Zhang, Xingyu and Hu, Yang and Jiang, Bo and He, Tong and Zhang, Qian and Long, Xiaoxiao and Yin, Wei},
  booktitle=CVPR,
  pages={1602--1611},
  year={2025}
}

@article{li2025driver1,
  title={Drive-R1: Bridging Reasoning and Planning in VLMs for Autonomous Driving with Reinforcement Learning},
  author={Li, Yue and Tian, Meng and Zhu, Dechang and Zhu, Jiangtong and Lin, Zhenyu and Xiong, Zhiwei and Zhao, Xinhai},
  journal={arXiv preprint arXiv:2506.18234},
  year={2025}
}

@article{jiao2025evadrive,
  title={Evadrive: Evolutionary adversarial policy optimization for end-to-end autonomous driving},
  author={Jiao, Siwen and Qian, Kangan and Ye, Hao and Zhong, Yang and Luo, Ziang and Jiang, Sicong and Huang, Zilin and Fang, Yangyi and Miao, Jinyu and Fu, Zheng and others},
  journal={arXiv preprint arXiv:2508.09158},
  year={2025}
}

@article{li2025recogdrive,
  title={Recogdrive: A reinforced cognitive framework for end-to-end autonomous driving},
  author={Li, Yongkang and Xiong, Kaixin and Guo, Xiangyu and Li, Fang and Yan, Sixu and Xu, Gangwei and Zhou, Lijun and Chen, Long and Sun, Haiyang and Wang, Bing and others},
  journal={arXiv preprint arXiv:2506.08052},
  year={2025}
}

@article{chi2025impromptu,
  title={Impromptu VLA: Open Weights and Open Data for Driving Vision-Language-Action Models},
  author={Chi, Haohan and Gao, Huan-ang and Liu, Ziming and Liu, Jianing and Liu, Chenyu and Li, Jinwei and Yang, Kaisen and Yu, Yangcheng and Wang, Zeda and Li, Wenyi and others},
  journal={arXiv preprint arXiv:2505.23757},
  year={2025}
}

@article{zhou2025opendrivevla,
  title={Opendrivevla: Towards end-to-end autonomous driving with large vision language action model},
  author={Zhou, Xingcheng and Han, Xuyuan and Yang, Feng and Ma, Yunpu and Tresp, Volker and Knoll, Alois},
  journal={arXiv preprint arXiv:2503.23463},
  year={2025}
}

@article{lipman2022flow,
  title={Flow matching for generative modeling},
  author={Lipman, Yaron and Chen, Ricky TQ and Ben-Hamu, Heli and Nickel, Maximilian and Le, Matt},
  journal={arXiv preprint arXiv:2210.02747},
  year={2022}
}

@article{chi2025diffusion,
  title={Diffusion policy: Visuomotor policy learning via action diffusion},
  author={Chi, Cheng and Xu, Zhenjia and Feng, Siyuan and Cousineau, Eric and Du, Yilun and Burchfiel, Benjamin and Tedrake, Russ and Song, Shuran},
  journal={The International Journal of Robotics Research},
  volume={44},
  number={10-11},
  pages={1684--1704},
  year={2025}
}

@article{guan2024world,
  title={World models for autonomous driving: An initial survey},
  author={Guan, Yanchen and Liao, Haicheng and Li, Zhenning and Hu, Jia and Yuan, Runze and Zhang, Guohui and Xu, Chengzhong},
  journal=TIV,
  year={2024},
  publisher={IEEE}
}

@article{kong20253d,
  title={3D and 4D world modeling: A survey},
  author={Kong, Lingdong and Yang, Wesley and Mei, Jianbiao and Liu, Youquan and Liang, Ao and Zhu, Dekai and Lu, Dongyue and Yin, Wei and Hu, Xiaotao and Jia, Mingkai and others},
  journal={arXiv preprint arXiv:2509.07996},
  year={2025}
}

@inproceedings{wang2024drivingfuture,
  title={Driving into the future: Multiview visual forecasting and planning with world model for autonomous driving},
  author={Wang, Yuqi and He, Jiawei and Fan, Lue and Li, Hongxin and Chen, Yuntao and Zhang, Zhaoxiang},
  booktitle=CVPR,
  pages={14749--14759},
  year={2024}
}

@inproceedings{zheng2025world4drive,
  title={World4drive: End-to-end autonomous driving via intention-aware physical latent world model},
  author={Zheng, Yupeng and Yang, Pengxuan and Xing, Zebin and Zhang, Qichao and Zheng, Yuhang and Gao, Yinfeng and Li, Pengfei and Zhang, Teng and Xia, Zhongpu and Jia, Peng and others},
  booktitle=ICCV,
  pages={28632--28642},
  year={2025}
}

@article{li2024law,
  title={Enhancing end-to-end autonomous driving with latent world model},
  author={Li, Yingyan and Fan, Lue and He, Jiawei and Wang, Yuqi and Chen, Yuntao and Zhang, Zhaoxiang and Tan, Tieniu},
  journal={arXiv preprint arXiv:2406.08481},
  year={2024}
}

@article{zhang2025epona,
  title={Epona: Autoregressive Diffusion World Model for Autonomous Driving},
  author={Zhang, Kaiwen and Tang, Zhenyu and Hu, Xiaotao and Pan, Xingang and Guo, Xiaoyang and Liu, Yuan and Huang, Jingwei and Yuan, Li and Zhang, Qian and Long, Xiao-Xiao and others},
  journal={arXiv preprint arXiv:2506.24113},
  year={2025}
}

@inproceedings{zhao2025drivedreamer4d,
  title={Drivedreamer4d: World models are effective data machines for 4d driving scene representation},
  author={Zhao, Guosheng and Ni, Chaojun and Wang, Xiaofeng and Zhu, Zheng and Zhang, Xueyang and Wang, Yida and Huang, Guan and Chen, Xinze and Wang, Boyuan and Zhang, Youyi and others},
  booktitle=CVPR,
  pages={12015--12026},
  year={2025}
}

@article{wang2025uniocc,
  title={Uniocc: A unified benchmark for occupancy forecasting and prediction in autonomous driving},
  author={Wang, Yuping and Huang, Xiangyu and Sun, Xiaokang and Yan, Mingxuan and Xing, Shuo and Tu, Zhengzhong and Li, Jiachen},
  journal={arXiv preprint arXiv:2503.24381},
  year={2025}
}

@inproceedings{zheng2024occworld,
  title={Occworld: Learning a 3d occupancy world model for autonomous driving},
  author={Zheng, Wenzhao and Chen, Weiliang and Huang, Yuanhui and Zhang, Borui and Duan, Yueqi and Lu, Jiwen},
  booktitle=ECCV,
  pages={55--72},
  year={2024},
  organization={Springer}
}

@article{li2025wote,
  title={End-to-end driving with online trajectory evaluation via bev world model},
  author={Li, Yingyan and Wang, Yuqi and Liu, Yang and He, Jiawei and Fan, Lue and Zhang, Zhaoxiang},
  journal={arXiv preprint arXiv:2504.01941},
  year={2025}
}

@article{li2024ssr,
  title={Navigation-guided sparse scene representation for end-to-end autonomous driving},
  author={Li, Peidong and Cui, Dixiao},
  journal={arXiv preprint arXiv:2409.18341},
  year={2024}
}

@article{yang2025worldrft,
  title={WorldRFT: Latent World Model Planning with Reinforcement Fine-Tuning for Autonomous Driving},
  author={Yang, Pengxuan and Lu, Ben and Xia, Zhongpu and Han, Chao and Gao, Yinfeng and Zhang, Teng and Zhan, Kun and Lang, XianPeng and Zheng, Yupeng and Zhang, Qichao},
  journal={arXiv preprint arXiv:2512.19133},
  year={2025}
}

@article{chen2024end,
  title={End-to-end autonomous driving: Challenges and frontiers},
  author={Chen, Li and Wu, Penghao and Chitta, Kashyap and Jaeger, Bernhard and Geiger, Andreas and Li, Hongyang},
  journal=PAMI,
  year={2024},
  publisher={IEEE}
}

@article{chib2023recent,
  title={Recent advancements in end-to-end autonomous driving using deep learning: A survey},
  author={Chib, Pranav Singh and Singh, Pravendra},
  journal=TIV,
  volume={9},
  number={1},
  pages={103--118},
  year={2023},
  publisher={IEEE}
}

@article{dong2025end,
  title={End-to-End Autonomous Driving: From Classic Paradigm to Large Model Empowerment—A Comprehensive Survey},
  author={Dong, Wei and Lu, Sikai and Chen, Xinhe and Zhang, Shunyao and Liu, Qingchao and Liu, Ze and Chen, Long and Wang, Hai and Cai, Yingfeng},
  journal={IEEE Internet of Things Journal},
  volume={13},
  number={3},
  pages={3870--3898},
  year={2025},
  publisher={IEEE}
}

@article{hu2025vision,
  title={Vision-language-action models for autonomous driving: Past, present, and future},
  author={Hu, Tianshuai and Liu, Xiaolu and Wang, Song and Zhu, Yiyao and Liang, Ao and Kong, Lingdong and Zhao, Guoyang and Gong, Zeying and Cen, Jun and Huang, Zhiyu and others},
  journal={arXiv preprint arXiv:2512.16760},
  year={2025}
}

@inproceedings{wang2023exploring,
  title={Exploring object-centric temporal modeling for efficient multi-view 3d object detection},
  author={Wang, Shihao and Liu, Yingfei and Wang, Tiancai and Li, Ying and Zhang, Xiangyu},
  booktitle=ICCV,
  pages={3621--3631},
  year={2023}
}

@inproceedings{song2020pip,
  title={Pip: Planning-informed trajectory prediction for autonomous driving},
  author={Song, Haoran and Ding, Wenchao and Chen, Yuxuan and Shen, Shaojie and Wang, Michael Yu and Chen, Qifeng},
  booktitle=ECCV,
  pages={598--614},
  year={2020},
  organization={Springer}
}

@inproceedings{xin2018intention,
  title={Intention-aware long horizon trajectory prediction of surrounding vehicles using dual LSTM networks},
  author={Xin, Long and Wang, Pin and Chan, Ching-Yao and Chen, Jianyu and Li, Shengbo Eben and Cheng, Bo},
  booktitle={ITSC},
  pages={1441--1446},
  year={2018},
  organization={IEEE}
}

@inproceedings{liu2021survey,
  title={A survey on deep-learning approaches for vehicle trajectory prediction in autonomous driving},
  author={Liu, Jianbang and Mao, Xinyu and Fang, Yuqi and Zhu, Delong and Meng, Max Q-H},
  booktitle={ROBIO},
  pages={978--985},
  year={2021},
  organization={IEEE}
}

@article{gonzalez2015review,
  title={A review of motion planning techniques for automated vehicles},
  author={Gonz{\'a}lez, David and P{\'e}rez, Joshu{\'e} and Milan{\'e}s, Vicente and Nashashibi, Fawzi},
  journal={TITS},
  volume={17},
  number={4},
  pages={1135--1145},
  year={2015},
  publisher={IEEE}
}

@article{lim2019hybrid,
  title={Hybrid trajectory planning for autonomous driving in on-road dynamic scenarios},
  author={Lim, Wonteak and Lee, Seongjin and Sunwoo, Myoungho and Jo, Kichun},
  journal={TITS},
  volume={22},
  number={1},
  pages={341--355},
  year={2019},
  publisher={IEEE}
}

@inproceedings{wang2024drivedreamer,
  title={Drivedreamer: Towards real-world-drive world models for autonomous driving},
  author={Wang, Xiaofeng and Zhu, Zheng and Huang, Guan and Chen, Xinze and Zhu, Jiagang and Lu, Jiwen},
  booktitle=ECCV,
  pages={55--72},
  year={2024}
 
}

@inproceedings{caesar2020nuscenes,
  title={nuscenes: A multimodal dataset for autonomous driving},
  author={Caesar, Holger and Bankiti, Varun and Lang, Alex H and Vora, Sourabh and Liong, Venice Erin and Xu, Qiang and Krishnan, Anush and Pan, Yu and Baldan, Giancarlo and Beijbom, Oscar},
  booktitle=CVPR,
  pages={11621--11631},
  year={2020}
}

@inproceedings{dauner2024navsim,
  title={Navsim: Data-driven non-reactive autonomous vehicle simulation and benchmarking},
  author={Dauner, Daniel and Hallgarten, Marcel and Li, Tianyu and Weng, Xinshuo and Huang, Zhiyu and Yang, Zetong and Li, Hongyang and Gilitschenski, Igor and Ivanovic, Boris and Pavone, Marco and others},
  booktitle=NeurIPS,
  year={2024}
}

@inproceedings{yang2025drivingoccworld,
  title={Driving in the occupancy world: Vision-centric 4d occupancy forecasting and planning via world models for autonomous driving},
  author={Yang, Yu and Mei, Jianbiao and Ma, Yukai and Du, Siliang and Chen, Wenqing and Qian, Yijie and Feng, Yuxiang and Liu, Yong},
  booktitle=AAAI,
  volume={39},
  pages={9327--9335},
  year={2025}
}

@inproceedings{podell2023sdxl,
  title={Sdxl: Improving latent diffusion models for high-resolution image synthesis},
  author={Podell, Dustin and English, Zion and Lacey, Kyle and Blattmann, Andreas and Dockhorn, Tim and M{\"u}ller, Jonas and Penna, Joe and Rombach, Robin},
  booktitle=ICLR,
  year={2024}
}

@inproceedings{gao2023magicdrive,
  title={Magicdrive: Street view generation with diverse 3d geometry control},
  author={Gao, Ruiyuan and Chen, Kai and Xie, Enze and Hong, Lanqing and Li, Zhenguo and Yeung, Dit-Yan and Xu, Qiang},
  booktitle=ICLR,
  year={2024}
}

@article{hu2023gaia,
  title={Gaia-1: A generative world model for autonomous driving},
  author={Hu, Anthony and Russell, Lloyd and Yeo, Hudson and Murez, Zak and Fedoseev, George and Kendall, Alex and Shotton, Jamie and Corrado, Gianluca},
  journal={arXiv preprint arXiv:2309.17080},
  year={2023}
}

@article{loshchilov2017decoupled,
  title={Decoupled weight decay regularization},
  author={Loshchilov, Ilya and Hutter, Frank},
  journal={arXiv preprint arXiv:1711.05101},
  year={2017}
}

@inproceedings{he2016deep,
  title={Deep residual learning for image recognition},
  author={He, Kaiming and Zhang, Xiangyu and Ren, Shaoqing and Sun, Jian},
  booktitle=CVPR,
  pages={770--778},
  year={2016}
}

@article{chen2024vadv2,
  title={Vadv2: End-to-end vectorized autonomous driving via probabilistic planning},
  author={Chen, Shaoyu and Jiang, Bo and Gao, Hao and Liao, Bencheng and Xu, Qing and Zhang, Qian and Huang, Chang and Liu, Wenyu and Wang, Xinggang},
  journal={arXiv preprint arXiv:2402.13243},
  year={2024}
}

@inproceedings{prakash2021multi,
  title={Multi-modal fusion transformer for end-to-end autonomous driving},
  author={Prakash, Aditya and Chitta, Kashyap and Geiger, Andreas},
  booktitle=CVPR,
  pages={7077--7087},
  year={2021}
}

@article{chitta2022transfuser,
  title={Transfuser: Imitation with transformer-based sensor fusion for autonomous driving},
  author={Chitta, Kashyap and Prakash, Aditya and Jaeger, Bernhard and Yu, Zehao and Renz, Katrin and Geiger, Andreas},
  journal=PAMI,
  volume={45},
  number={11},
  pages={12878--12895},
  year={2022},
  publisher={IEEE}
}

@inproceedings{weng2024paradrive,
  title={Para-drive: Parallelized architecture for real-time autonomous driving},
  author={Weng, Xinshuo and Ivanovic, Boris and Wang, Yan and Wang, Yue and Pavone, Marco},
  booktitle=CVPR,
  pages={15449--15458},
  year={2024}
}

@article{yuan2024drama,
  title={Drama: An efficient end-to-end motion planner for autonomous driving with mamba},
  author={Yuan, Chengran and Zhang, Zhanqi and Sun, Jiawei and Sun, Shuo and Huang, Zefan and Lee, Christina Dao Wen and Li, Dongen and Han, Yuhang and Wong, Anthony and Tee, Keng Peng and others},
  journal={arXiv preprint arXiv:2408.03601},
  year={2024}
}

@article{li2024hydra,
  title={Hydra-mdp: End-to-end multimodal planning with multi-target hydra-distillation},
  author={Li, Zhenxin and Li, Kailin and Wang, Shihao and Lan, Shiyi and Yu, Zhiding and Ji, Yishen and Li, Zhiqi and Zhu, Ziyue and Kautz, Jan and Wu, Zuxuan and others},
  journal={arXiv preprint arXiv:2406.06978},
  year={2024}
}

@inproceedings{song2025momoad,
  title={Don't shake the wheel: Momentum-aware planning in end-to-end autonomous driving},
  author={Song, Ziying and Jia, Caiyan and Liu, Lin and Pan, Hongyu and Zhang, Yongchang and Wang, Junming and Zhang, Xingyu and Xu, Shaoqing and Yang, Lei and Luo, Yadan},
  booktitle=CVPR,
  pages={22432--22441},
  year={2025}
}

@inproceedings{allen1983planning,
  title={Planning using a temporal world model},
  author={Allen, James F and Koomen, Johannes A},
  booktitle=IJCAI,
  pages={741--747},
  year={1983}
}

@inproceedings{liu2021swin,
  title={Swin transformer: Hierarchical vision transformer using shifted windows},
  author={Liu, Ze and Lin, Yutong and Cao, Yue and Hu, Han and Wei, Yixuan and Zhang, Zheng and Lin, Stephen and Guo, Baining},
  booktitle=ICCV,
  pages={10012--10022},
  year={2021}
}

@article{liu2022flow,
  title={Flow straight and fast: Learning to generate and transfer data with rectified flow},
  author={Liu, Xingchao and Gong, Chengyue and Liu, Qiang},
  journal={arXiv preprint arXiv:2209.03003},
  year={2022}
}

\end{document}